\documentclass{article}
\usepackage{arxiv}
\usepackage[utf8]{inputenc}
\usepackage{xcolor}
\usepackage[authoryear,round]{natbib}
\setcitestyle{citesep={;},aysep={,},yysep={;}}
\renewcommand{\shorttitle}{Decoding Naturalistic Speech from fMRI}

\usepackage{amsmath,amsfonts,bm}
\newcommand{\ve}{{\bm{e}}}
\newcommand{\vy}{{\bm{y}}}
\newcommand{\mX}{{\bm{X}}}

\usepackage{hyperref}
\usepackage{cleveref}
\usepackage{url}
\usepackage{graphicx}
\usepackage{float}
\usepackage{booktabs}
\usepackage{amssymb}
\usepackage{listings}
\usepackage{inconsolata}
\usepackage{multirow}
\usepackage{bbm}

\definecolor{blue600}{HTML}{2563eb}

\hypersetup{colorlinks=true, linkcolor=blue600, citecolor=blue600, urlcolor=blue600}

\title{Decoding Naturalistic Speech from fMRI with Temporal Modeling and Multimodal Alignment}

\author{%
\textbf{Corentin Caris$^{1,}$\thanks{Equal contribution.}\,\,, Louis Jalouzot$^{1,2,3,*}$, Nona Rajabi$^{1,*}$, Géraud Ilinca$^{1}$,}\\
\textbf{Ahmed Hassayoune$^{1}$, Asilbek Muminov$^{1}$, Thaddée Delebarre$^{1}$, Clara Fonteneau$^{1}$,}\\
\textbf{Charles Maussion$^{1}$, Bertrand Thirion$^{1,4}$, Alexis Thual$^{1}$, Antoine Collas$^{1}$}\\[0.5em]
$^{1}$Karavela\\
$^{2}$UNICOG, CNRS, INSERM, CEA, Paris-Saclay University\\
$^{3}$LSCP, EHESS, ENS, CNRS, PSL University\\
$^{4}$Inria, CEA, Paris-Saclay University
}

\begin{document}

\maketitle

\begin{abstract}
Reconstructing what a person perceives from their brain activity is a stringent test of whether neural recordings carry meaningful information.
Naturalistic speech offers a rich testbed for this challenge: it unfolds over multiple timescales, but fMRI's noisy, temporally smeared measurements make this structure hard to recover.
We introduce NOEMA, a temporal convolutional decoder that aligns windows of consecutive fMRI measurements with multimodal stimulus embeddings.
These embeddings combine complementary text, audio, and speech representations of the same stimulus segment, enabling its retrieval directly from brain activity.
On two naturalistic speech datasets, \textit{Cephalonauts One} and \textit{LeBel}, NOEMA retrieves the correct segment among 2,000 candidates within the top 10 for more than 42\% of held-out queries.
This substantially outperforms baseline decoders, an advantage that persists when the candidate pool is extended with thousands of distractors absent from training.
Controlled ablations establish the contributions of temporal modeling and complementary stimulus representations.
\end{abstract}

\section{Introduction}

Can one gain access to the cognitive experience of a living being from recordings of their brain activity?
A growing body of work shows that machine learning models can reconstruct what a participant is viewing, hearing, or even imagining from brain activity alone \citep{horikawaGenericDecodingSeen2017}.
This \textit{decoding} endeavor is tightly linked to representation learning: if a model can precisely predict cognitive states from brain activity, it should have extracted meaningful structures from the signals at hand, opening a window into a better understanding of the complexity of the living brain.

Seminal work from \cite{haxbyDistributedOverlappingRepresentations2001} showed that fMRI measurements of the ventral visual stream contain sufficient information to distinguish the category of a viewed image from a small set of categories.
Recent data collection efforts like the \textit{Natural Scenes Dataset} \citep{allenMassive7TFMRI2022} have since scaled up the number of image categories, enabling more accurate decoders \citep{ozcelikNaturalSceneReconstruction2023}.
Yet, since these stimuli are static, these decoders mostly rely on spatial information - where the activations are happening on the cortex - to perform their task.
Given the inherently dynamic nature of the brain, it is crucial to develop decoding approaches that can leverage, and therefore reveal, the temporal structure of the underlying neural processes.

\begin{figure}[t]
    \centering
    \includegraphics[width=\linewidth]{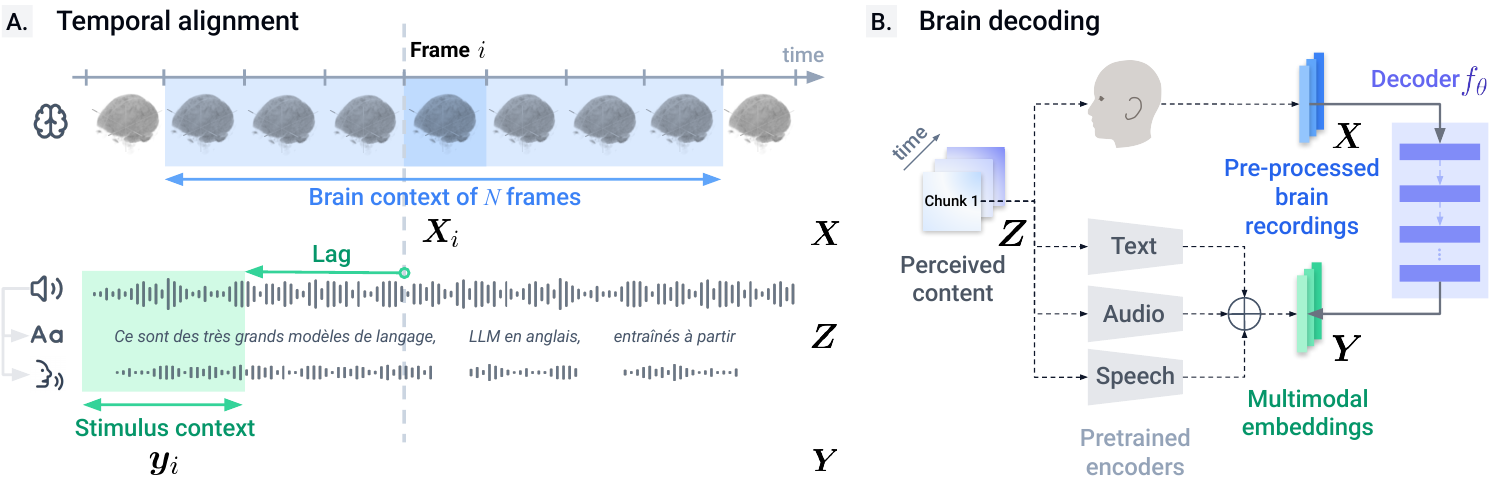}
    \caption{\textbf{Temporal alignment and stimulus decoding.}
    \textbf{A.} A temporal offset (lag) aligns each stimulus-context endpoint with the central fMRI frame to account for the hemodynamic delay.
    \textbf{B.} A decoder maps brain recordings to vector representations of segments produced by pretrained stimulus encoders.
    Timing values are schematic.}
    \label{fig:decoding_framework}
\end{figure}

To this end, naturalistic speech stimuli are particularly promising: unlike static images, they unfold over time, giving decoders temporal structure to exploit.
Yet fMRI trades temporal precision for spatial resolution, while M/EEG offer the reverse trade-off.
\cite{pereira2018universal} showed that fMRI can reconstruct the semantic content of words and sentences, primarily by leveraging spatial patterns of brain activity, while \cite{defossez_decoding_2023} showed that M/EEG support speech decoding through their finer temporal dynamics.
This suggests that spatial and temporal information may offer complementary routes to decoding naturalistic speech.
Seeking to leverage both, we introduce Neural Ordered Embeddings with Multimodal Alignment (NOEMA), a temporal convolutional decoder that maps windows of consecutive fMRI measurements to multimodal representations of the corresponding speech segments (Figure~\ref{fig:decoding_framework}).
On two large, densely-sampled speech datasets, NOEMA substantially outperforms existing decoding baselines on held-out sessions and extended retrieval sets.

\section{Related work}
\label{sec:related_work}

\textbf{Natural language decoding from fMRI.}
\cite{pereira2018universal} trained semantic decoders on individual concepts and generalized to sentences, while Brain2Word \citep{affolter2020brain2word} decoded words from a fixed vocabulary.
For naturalistic listening, \cite{huth_natural_2016} modeled cortical responses to semantic features; \cite{tang_semantic_2023} combined an encoding model with language-model-guided beam search to reconstruct continuous text.
BrainLLM \citep{ye2025generative} instead feeds fMRI-derived embeddings directly into a language model.

\textbf{Temporal integration of fMRI frames.}
Prior methods integrate temporal context in different ways.
\cite{jalouzot_optimizing_2025} average consecutive volumes before decoding text embeddings.
\emph{UniCoRN} \citep{xi2023unicorn} combines snapshot and sequence reconstruction pretraining with text generation.
BP-GPT \citep{chen2025bp} encodes fMRI windows into language-model prompts, using contrastive alignment with text prompts alongside a generation loss.
\emph{CogReader} \citep{lu2025cogreader} combines spatiotemporal encoding with semantic summaries carried between generated segments.
UniCoRN reports lower generation scores under stimulus-disjoint than random splits; BP-GPT evaluates on a held-out story, and CogReader uses stimulus-disjoint splits from naturalistic stories.
These generation results are not directly comparable to segment retrieval.
Here, we compare learned temporal convolutions with frame-wise decoders and an MLP receiving the same context.

\textbf{Cross-modal alignment with contrastive learning.}
CLIP \citep{radford2021learning} aligns paired modalities for retrieval in a shared space.
\cite{defossez_decoding_2023} aligned MEG and EEG with heard-audio representations, while MindEye \citep{scotti_reconstructing_2023} used contrastive alignment with CLIP image representations for fMRI decoding.
Language-model representations predict neural responses to language \citep{schrimpf2021neural,caucheteux_brains_2022,pasquiou2022neural}, and can guide stimulus selection to modulate language-network responses \citep{tuckuteDrivingSuppressingHuman2024}.
Beyond text, \textit{Cephalonauts One} \citep{collas2026cephalonauts} combines text and audio targets for single-frame decoding.
Distinct cortical contributions of acoustic, articulatory, and semantic features \citep{deheer2017hierarchical} motivate our evaluation of complementary targets with temporal decoding.

\textbf{Contributions.}
We introduce NOEMA, a temporal convolutional decoder mapping fMRI windows to multimodal speech representations.
Our experiments show that NOEMA (1) outperforms the evaluated frame-wise baselines and a Context MLP receiving the same seven-frame input on two naturalistic listening datasets with held-out sessions and unseen stimuli, exceeding 42\% top-10 accuracy among approximately 2,000 candidates; (2) benefits from temporal context and frame order, as tested by context-length and shuffle ablations; (3) retains its retrieval advantage when candidates from unseen episodes of the same podcast are added; and (4) improves with complementary text, audio, and speech representations.

\begin{figure}[t]
    \centering
    \includegraphics[width=\linewidth]{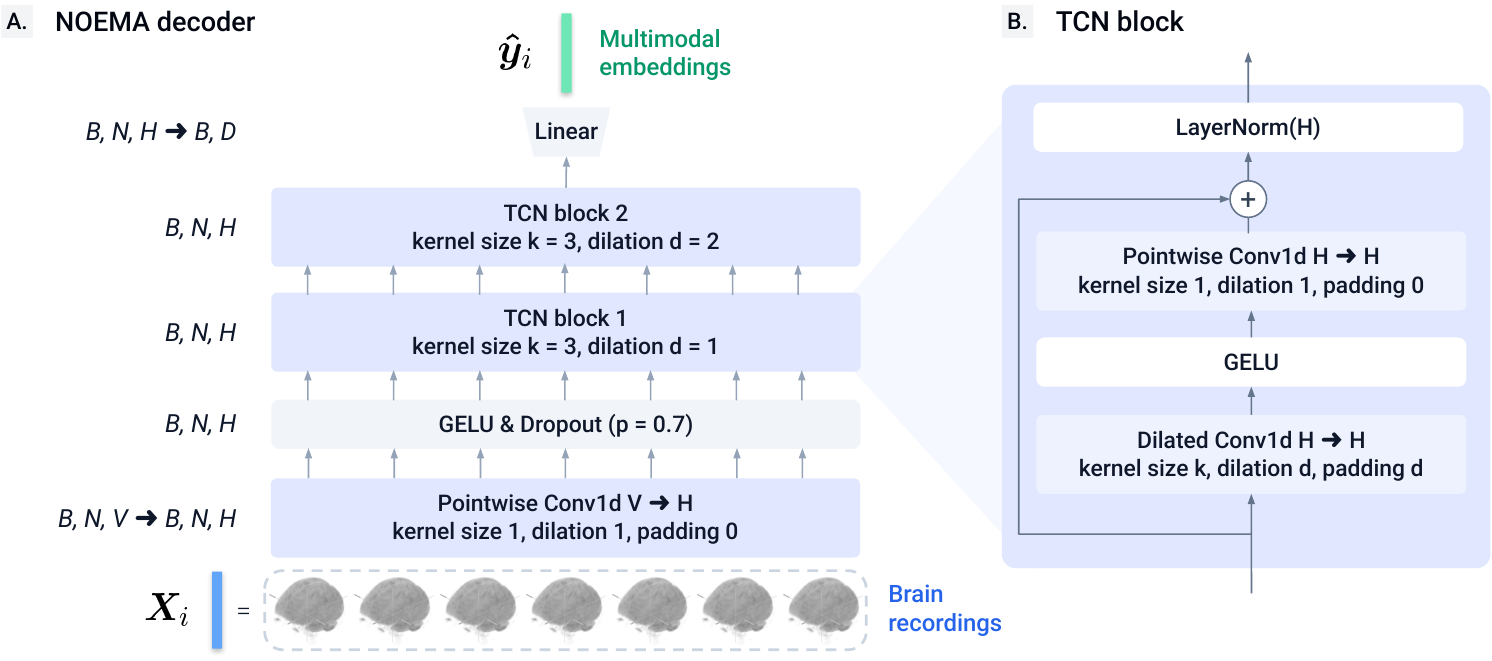}
    \caption{\textbf{NOEMA architecture.}
    \textbf{A.} Information flow from a window of brain recordings to the predicted multimodal embedding. Tensor dimensions are shown beside each stage; only the central-frame representation is projected into the multimodal embedding space.
    B, N, V, D, H denote the batch size, number of frames, cortical features, target dimension, and hidden dimension respectively.
    \textbf{B.} Structure of a residual temporal convolutional block (TCN block), with a skip connection that bypasses both convolutions and adds the block input before layer normalization. These convolutions use stride one, include biases, and preserve the hidden dimension; symmetric padding preserves the temporal length.}
    \label{fig:noema_architecture}
\end{figure}

\section{Method}
\label{sec:method}

\subsection{Problem formulation and retrieval objective}
\textbf{Decoding task.}
We frame naturalistic speech decoding as a retrieval problem: given fMRI measurements, the decoder must identify the corresponding perceived speech segment among candidate segments.
As illustrated in Figure~\ref{fig:decoding_framework}, we learn a participant-specific decoder $f_\theta$ that maps $\mX_i\in\mathbb{R}^{N\times V}$, a window of $N$ fMRI frames with $V$ cortical features, to $\vy_i\in\mathbb{R}^{D}$, the embedding of the corresponding speech segment.
At evaluation, we rank the cosine similarities $s_{ij}$ between the predicted embedding $\hat{\vy}_i \triangleq f_\theta(\mX_i)$ and each of $M$ candidate segment embeddings $\vy_j$ from a held-out session.
In particular, at inference, the decoder receives only fMRI; candidate audio and transcripts are used solely to precompute the fixed stimulus embeddings.

\textbf{Contrastive alignment.}
We train the fMRI decoder by aligning its predictions with the multimodal stimulus targets
using the CLIP \citep{radford2021learning} contrastive loss defined over a batch of $B$ samples:
\begin{equation*}
    \mathcal{L}_{\text{CLIP}} \quad \triangleq \quad -\frac{1}{2B} \; \sum_{i=1}^{B} \;
    \left[ \;
    \log\frac{\exp(s_{ii}/\tau)}{\sum_{j=1}^{B}\exp(s_{ij}/\tau)}
    \; + \;
    \log\frac{\exp(s_{ii}/\tau)}{\sum_{j=1}^{B}\exp(s_{ji}/\tau)}
    \; \right]
\end{equation*}
In this loss, $i,j$ index paired examples within the batch, and $s_{ii}$ scores a matched brain--stimulus pair. The temperature $\tau>0$ is optimized per model as described in Section~\ref{sec:training}.

\subsection{NOEMA Architecture}
\label{sec:architecture}

\textbf{Temporal window.}
NOEMA instantiates $f_\theta$ as a temporal convolutional decoder that maps $N=7$ consecutive fMRI frames, each lasting $\Delta$ seconds, to the stimulus embedding associated with the central frame (Figure~\ref{fig:noema_architecture}A).
For a given stimulus context ending at time $t$, the associated fMRI window is centered at $t+\delta$ (where $\delta=2.5$ seconds), and lasts $N\times\Delta$ seconds (see Section~\ref{sec:datasets} for values).
We use an alignment offset of 2.5 s, which yields the highest accuracy for the MLP and near-peak accuracy for NOEMA (Appendix~\ref{app:lag_ablation}).

\textbf{Spatial projection.}
A pointwise (1$\times$1) convolution maps $V$ cortical features to a hidden width of $H=4096$, followed by GELU and dropout with probability $0.7$.
More simply, this convolution applies the same linear projection independently to each frame, preserving temporal order and yielding an $N\times H$ representation per sample.

\textbf{Temporal mixing.}
Two residual temporal convolutional blocks \citep{bai2018empirical} integrate consecutive frames using kernel size $K=3$ and dilations $d_1=1$ and $d_2=2$, measured in frames.
Each TCN block applies a dilated convolution, GELU, and a pointwise convolution, followed by residual addition and layer normalization (Figure~\ref{fig:noema_architecture}B).
Together, the blocks give the central representation a seven-frame receptive field that includes both earlier and later measurements, making the decoder noncausal.

\textbf{Central-frame prediction.}
A final linear layer maps the central representation of dimension $H$ to the $D=8704$-dimensional multimodal target space (Section~\ref{sec:stimuli}).
Only this prediction contributes to the training loss and retrieval score.
Appendix~\ref{app:noema_architecture} summarizes the decoder architecture.

\subsection{Multimodal stimulus representations}
\label{sec:stimuli}
\textbf{Stimulus encoders.}
We construct multimodal stimulus targets from three frozen encoders operating on
text, audio, and speech, extracted at a sequence of time markers $t$ spaced by $\Delta$
seconds. Text and audio share a longer context $(t-T,t]$ with $T=10$ seconds: transcript
words are selected by their end times, and audio uses the corresponding waveform segment.
\texttt{Qwen3-Embedding-8B}~\citep{zhang_qwen3_2025} embeds the transcripts and
supplies 4096 text features from the last non-padding token of its final hidden layer.
\texttt{LCO-Embedding-Omni-7B}~\citep{xiao2025scaling} supplies 3584 audio
features from the audio waveform.
Finally, the ASR encoder \texttt{Parakeet-TDT-0.6B-v3}~\citep{sekoyan2025parakeet}
 uses a window of size $\Delta$ seconds, $(t-\Delta,t]$, producing 1024 speech features by averaging
its penultimate encoder-layer states within that interval. This shorter context provides
a temporally precise speech target that complements the broader text and audio contexts.
Since this window matches the marker spacing, successive speech targets tile the stimulus
without overlap, giving them a sequential, window-specific organization. We evaluate the effect
of changing the speech context in Appendix~\ref{app:asr_context_ablation}.
The encoders are used only to precompute targets for training and evaluation; their parameters remain fixed while the participant-specific fMRI decoder is trained.

\textbf{Target embedding.}
We $\ell_2$-normalize each encoder output $\ve_i^{\mathrm{text}}$, $\ve_i^{\mathrm{audio}}$, and $\ve_i^{\mathrm{speech}}$, then concatenate them into a $8704$-dimensional target:
\begin{equation*}
    \vy_i \quad \triangleq \quad \left[ \;
    \frac{\ve_i^{\mathrm{text}}}{\|\ve_i^{\mathrm{text}}\|_2} \; ; \;\,
    \frac{\ve_i^{\mathrm{audio}}}{\|\ve_i^{\mathrm{audio}}\|_2} \; ; \; \,
    \frac{\ve_i^{\mathrm{speech}}}{\|\ve_i^{\mathrm{speech}}\|_2}
    \; \right]
\end{equation*}
Each modality is normalized separately before concatenation; cosine similarity then normalizes the complete predicted and target vectors.

\section{Experimental setup}
\label{sec:experimental_setup}

\subsection{Datasets, preprocessing, and splits}
\label{sec:datasets}

\textbf{Datasets.}
We evaluate NOEMA on the naturalistic speech decoding task across two datasets, training and evaluating a separate decoder for each participant in each dataset independently:
\begin{itemize}
\item \textit{Cephalonauts One} \citep{collas2026cephalonauts}: fMRI recordings from three participants listening to podcasts in their native language, French. Each participant completed approximately 30 hours across 24 sessions; sessions are shared across participants, but no stimulus is repeated within a participant.
\item \textit{LeBel}~\citep{lebel_natural_2023}: fMRI recordings from eight participants listening to 6 hours of English stories. We focus on the three participants who completed an additional 10 hours, for sufficient training data, and exclude localizer runs and the repeatedly presented story \emph{Where There's Smoke}.
\end{itemize}

Table~\ref{tab:retained_data} in Appendix~\ref{app:retained_data} reports the number of sessions, runs, and acquisition hours for each participant.

\textbf{Preprocessing.}
For both datasets, we use DeepPrep~\citep{ren2025deepprep} to process fMRI images and reconstruct \texttt{fsaverage5} cortical surfaces, concatenate vertex data from both hemispheres, and detrend and standardize each run's vertex time series.
\textit{Cephalonauts One} recordings have a native repetition time (TR) of 0.52 seconds.
Following the Cephalonauts One preprocessing protocol~\citep{collas2026cephalonauts}, we apply a four-TR moving average and retain every fourth frame, yielding a frame spacing of $\Delta=2.08$ seconds.
\textit{LeBel}'s native TR already matches the stimulus marker spacing ($\Delta=2.0$ seconds), so we keep the original frame spacing without additional smoothing.
Using this preprocessed data, we form the seven-frame fMRI windows
described in Section~\ref{sec:architecture}.

\textbf{Dataset splits.}
Consecutive fMRI frames are temporally correlated and share stimulus context,
so random splits would leak information between training and evaluation.
We therefore split by recording session, which, given the datasets' design,
also ensures that no stimulus is shared across training, validation, and test sets.
Each model is evaluated on five splits, each holding out one session for testing,
one for validation, and using the rest for training.
For \textit{Cephalonauts One}, the five test sessions are those of the dataset benchmark~\citep{collas2026cephalonauts}.

\subsection{Baselines, training, and evaluation}
\label{sec:training}
\textbf{Baselines.}
We compare NOEMA with four baselines:
(1) \texttt{Ridge}, a center-frame Ridge regressor with L2 regularization;
(2) \texttt{MLP}, a single-frame MLP with one hidden layer and GELU activation, from \cite{collas2026cephalonauts};
(3) \texttt{MindEye MLP}, a deeper frame-wise MLP with four residual blocks and GELU activation, adapted from \cite{scotti_reconstructing_2023};
and (4) \texttt{Context MLP}, which applies \texttt{MLP}'s architecture to concatenated fMRI frames, controlling for access to the same temporal input window.
All baselines predict the same concatenated target embedding as NOEMA, using identical training and evaluation splits.
Unless stated otherwise, NOEMA and Context MLP receive seven-frame windows while the remaining baselines receive only the central frame; Context MLP keeps its name even when this window size is varied in the temporal-context ablation (Section~\ref{sec:results}), to distinguish it from the single-frame MLP.

\textbf{Training.}
The Ridge baseline is fitted in closed form using $\alpha=4.5\times10^{3}$,
selected by the hyperparameter search described in Appendix~\ref{app:hpo}.
All other baselines and NOEMA are trained with AdamW~\citep{loshchilov2017decoupled}
for up to 200 epochs with an early stopping patience of 10 epochs, using hyperparameters (learning rate, weight decay, batch size, temperature) tuned through a Bayesian search with Optuna~\citep{akiba2019optuna}.
For NOEMA, this yields a learning rate of $1.1\times10^{-4}$, a weight decay of $6.8\times10^{-3}$, a batch size of $4096$, and a temperature of $6.2\times10^{-3}$; the complete search space and final settings for every model are given in Appendix~\ref{app:hpo}.
Training uses a single NVIDIA H100 GPU (80GB) or AMD MI250X GPU.

\textbf{Retrieval settings.}
For model comparisons and ablations on both datasets, the retrieval set consists
of all stimulus embeddings from the held-out test session. The retrieval-set extension
experiment (Section~\ref{sec:open_dictionary_results}) instead evaluates generalization
beyond the original stimulus collection: for \textit{Cephalonauts One}, we progressively
augment each test-session retrieval set with up to 8,494 embeddings corresponding
to ten-second segments from 50 additional episodes of the same radio podcast, absent
from the original dataset, reaching a retrieval set roughly five times larger than the original at the largest extension.

\begin{figure}[t]
    \centering
    \includegraphics[width=\linewidth]{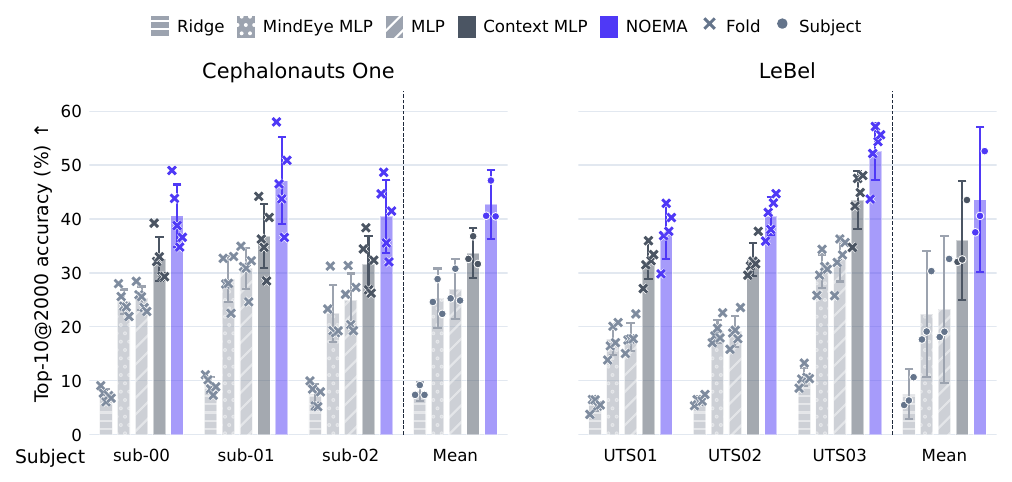}
    \caption{\textbf{Top-10@2000 stimulus retrieval across decoder configurations.}
Baselines are shown in light gray - with horizontal, dotted and diagonal hatchings for Ridge, MindEye MLP, and MLP respectively - Context MLP in dark gray, and NOEMA in indigo.
Subject columns show mean accuracy and standard deviation across five folds, with folds indicated by crosses.
In the Mean column, circles report subject means, while bars and error bars show the mean and 90\% Student-$t$ confidence interval across the three subject means.}
    \label{fig:retrieval_comparison}
\end{figure}

\textbf{Evaluation metrics.}
We evaluate retrieval by ranking $M$ candidate stimulus embeddings by cosine similarity to each predicted embedding.
Let $r_i$ denote the rank of the correct target for test sample $i$, with rank 1 corresponding to the highest similarity.
Our primary metric is Top-10 accuracy, the fraction of test samples for which the correct target appears among the ten highest-ranked candidates:
\begin{equation*}
\text{Top-10@}M = \frac{1}{n}\sum_{i=1}^{n}\mathbbm{1}[r_i\leq 10],
\end{equation*}
where $n$ is the number of test samples and $\mathbbm{1}[\cdot]$ is the indicator function.
The shorthand @2000 denotes approximately 2,000 candidates per test session.
To characterize retrieval beyond this fixed cutoff, we also report median rank and its normalized form, median relative rank:
\begin{equation*}
\text{Median rank} = \operatorname{median}(r_1,\ldots,r_n),
\qquad
\text{Median relative rank} = \frac{\text{Median rank}}{M}.
\end{equation*}
Lower values indicate better retrieval for both rank metrics.
We use median relative rank to compare retrieval sets of different sizes.

\section{Results}
\label{sec:results}

\subsection{Retrieval performance}
We first evaluate how accurately each decoder identifies the speech segment
that elicited a given neural response.
Figure~\ref{fig:retrieval_comparison} compares NOEMA with the baseline decoders
on \textit{Cephalonauts One} and \textit{LeBel} datasets. All models predict the same multimodal
(text, audio, speech) targets and are tuned according to the hyperparameter
optimization procedure described in Appendix~\ref{app:hpo}. NOEMA consistently
outperforms all baselines across subjects, retrieving the correct speech segment
within the top 10 candidates for 42.7\% of queries on \textit{Cephalonauts One} and 43.6\% on
\textit{LeBel}, far above the chance level of 0.5\%.
Notably, NOEMA also surpasses the Context MLP, even though both models receive
the same input context and the Context MLP has more parameters (Appendix~\ref{app:model_size}). NOEMA's
advantage therefore does not appear to only arise from additional context or
greater model capacity. One plausible explanation is its use of temporal
convolutions, which explicitly model the temporal structure of the neural
signal. This advantage holds across alternative retrieval metrics and rank
thresholds: Top-1@2000 accuracy, median relative rank, cumulative rank distributions,
and rank histograms are reported in Appendix~\ref{app:additional_results}.

\subsection{Stimulus representations and temporal context}

\begin{figure}[t]
    \centering
    \includegraphics[width=\linewidth]{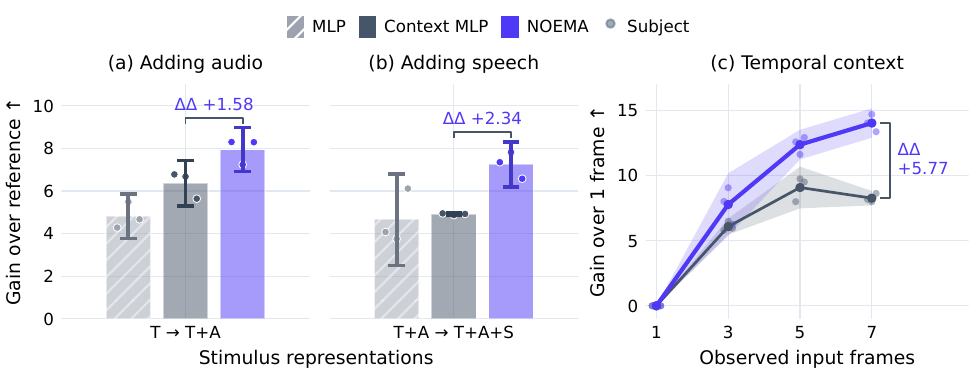}
\caption{\textbf{Effect of stimulus representations and temporal context (\textit{Cephalonauts One}).}
All panels show the gain in Top-10@2000 accuracy (percentage points) over a reference configuration, computed within each model, subject, and fold.
\textbf{(a)}~Adding audio to a text-only target (T$\rightarrow$T+A).
\textbf{(b)}~Adding speech to a text and audio target (T+A$\rightarrow$T+A+S).
T: Qwen text; A: LCO audio; S: Parakeet speech (ASR).
$\Delta\Delta$ is the mean NOEMA gain minus the mean Context MLP gain.
\textbf{(c)}~Gain from increasing the input from one frame to 3, 5, or 7 frames, for Context MLP and NOEMA.
Here, $\Delta\Delta$ compares the two models' gains from one to seven frames.
Circles show subject means over five folds.
Bars in (a, b) and lines in (c) show the mean across three subjects.
Error bars and shaded bands show 90\% Student-$t$ confidence intervals.
See Appendix~\ref{app:abs_stim_rep} and~\ref{app:abs_temporal_context} for absolute accuracies.
}
\label{fig:modality_temporal_ablation}
\end{figure}

\textbf{Stimulus representations.}
To assess the contribution of each stimulus representation, we trained NOEMA, MLP, and Context MLP on targets built from cumulative combinations of text (T), audio (A), and speech (S) representations.
On \textit{Cephalonauts One}, adding audio to a text-only target raises NOEMA's Top-10@2000 accuracy
from 27.6\% to 35.5\% (Figure~\ref{fig:modality_temporal_ablation}a), and further adding speech to the text-and-audio target raises it to 42.7\% (Figure~\ref{fig:modality_temporal_ablation}b).
Adding audio and then speech improves retrieval for all three models in every subject and fold, with the largest mean gains for NOEMA.
These gains exceed those of Context MLP, which receives the same seven-frame input, by 1.58 and 2.34 percentage points, respectively.
The same pattern holds on \textit{LeBel}, where adding speech has an even larger effect (Appendix~\ref{app:lebel_modality_temporal_ablation}).
Absolute accuracies for each model under the T, T+A, and T+A+S targets are reported in
Appendix~\ref{app:abs_stim_rep}.

\textbf{Temporal context.}
To assess how effectively each architecture exploits information spread across neighboring fMRI frames, we train Context MLP and NOEMA with 1, 3, 5, or 7 input frames.
Adding frames increases Context MLP's parameter count, while NOEMA's remains fixed.
Both models benefit from additional frames, with larger accuracy gains for NOEMA (Figure~\ref{fig:modality_temporal_ablation}c).
On \textit{Cephalonauts One}, NOEMA improves steadily with each added frame,
rising from 28.7\% Top-10@2000 accuracy with one frame to 42.7\% with seven. Context MLP, in contrast,
rises from 25.4\% to a peak of 34.5\% at five frames and declines slightly to
33.7\% at seven.
On \textit{LeBel}, both models improve at every window length, and again NOEMA benefits more
(Appendix~\ref{app:lebel_modality_temporal_ablation}).
NOEMA achieves this higher accuracy with fewer parameters: 253.8M trainable parameters, compared
with 623.0M for the seven-frame Context MLP (Appendix~\ref{app:model_size}).
These results suggest that NOEMA's temporal
convolutions make better use of additional fMRI frames than the tested Context MLP.
Absolute accuracies are reported in Appendix~\ref{app:abs_temporal_context}.
To probe the role of this temporal inductive bias, we retrain and evaluate
both models with their input frames permuted. A permutation shared across
samples can be offset by reordering the Context MLP's input weights, but it
disrupts the temporal neighborhoods on which NOEMA's convolutions rely.
Accordingly, NOEMA's accuracy drops while the Context MLP's remains
essentially unchanged, suggesting that NOEMA's inductive bias is well matched
to the chronological order of fMRI frames (Appendix~\ref{app:brain_context_shuffle}).

\subsection{Extended retrieval sets}
\label{sec:open_dictionary_results}

\begin{figure}[t]
    \centering
    \includegraphics[width=\linewidth]{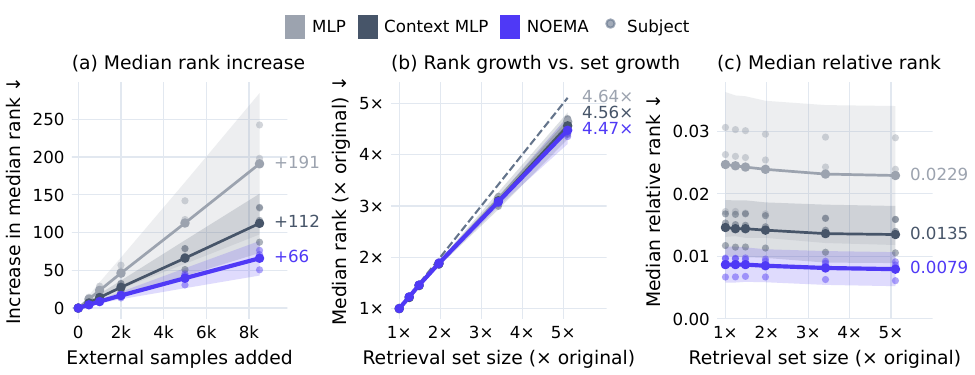}
    \caption{\textbf{Extended retrieval sets (\textit{Cephalonauts One}).}
Up to 8,494 segments from unseen episodes are added to each test-session retrieval set.
\textbf{(a)}~Increase in median rank versus number of added samples.
\textbf{(b)}~Median rank relative to the original set versus retrieval-set size; the dashed line marks proportional growth, expected if external samples were as confusable as in-set ones.
\textbf{(c)}~Median rank divided by retrieval-set size.
Lower is better.
Circles show subject means over five folds; lines and bands show the mean and 90\% Student-$t$ confidence interval across three subjects.}
    \label{fig:open_dictionary}
\end{figure}

\textbf{Retrieval beyond the original dataset.}
Retrieval within a single test session uses a closed set of about 2,000 candidates.
A practical decoder, however, must identify the correct segment within a much larger, open-ended pool.
To test whether our conclusions hold in this setting, we extend each \textit{Cephalonauts One} test-session retrieval set with up to 8,494 ten-second segments drawn from 50 episodes of the same radio podcast that are not part of the dataset (Appendix~\ref{app:extended_retrieval_set}).
Drawing distractors from the same show limits broad differences in recording conditions and stimulus domain;
this reduces, but does not eliminate, acoustic and stylistic differences between original and added candidates.
All models are evaluated without retraining, and we track the median rank of the correct segment as the retrieval set grows.
If the added distractors were exactly as confusable as the original candidates, the median rank would grow in proportion to the retrieval-set size, and the median relative rank would stay constant.

\textbf{Effect of retrieval-set size.}
NOEMA shows the smallest absolute increase in median rank as distractors are added.
At the largest extension, its median rank increases by 66 positions, compared with 112 and 191 for the
Context MLP and MLP baselines (Figure~\ref{fig:open_dictionary}a).
Relative to their original median rank, the ranks of all three models grow by a similar factor
(4.5-4.6$\times$ for a 5.1$\times$ larger set; Figure~\ref{fig:open_dictionary}b).
Because NOEMA already ranks the correct target within the top 0.9\% of candidates, compared with
2.5\% for the MLP, it moves fewer positions as the pool grows.
Across all three models, median rank increases slightly less than proportionally to retrieval-set size, yielding a small decrease in median relative rank (Figure~\ref{fig:open_dictionary}c).
NOEMA maintains the lowest median relative rank throughout the expansion.

\subsection{Model interpretation}
\label{sec:model_interpretation}

\begin{figure}[!t]
    \centering
    \includegraphics[width=\linewidth]{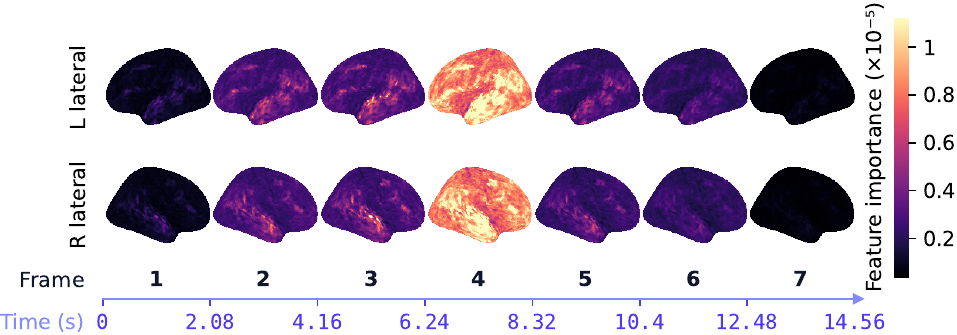}
    \caption{
        \textbf{Feature importance across input frames.}
        Absolute integrated-gradients attributions with respect to the 7 fMRI frames which NOEMA takes as input.
        The figure shows the left and right lateral views for \textit{Cephalonauts One}'s \texttt{sub-01}.
        Other views and more details on the method can be found in Appendix \ref{app:feature_importance}.
    }
    \label{fig:feature_importance}
\end{figure}

Finally, we examine the models' sensitivity to their fMRI inputs using feature-importance maps derived from integrated gradients~\citep{sundararajan2017axiomatic}.
\Cref{fig:feature_importance} shows NOEMA's left and right lateral views.
Details on the method, other views for NOEMA, and maps for the standard and Context MLP baselines can be found in Appendix~\ref{app:feature_importance}.
Among the seven-frame models, only NOEMA shows a bias toward the central frames which is consistent with its temporal convolutions and central-frame readout.
Across all three models, importance is higher in lateral temporal cortex, with smaller patches in frontal cortex, while the medial views show no distinct high-importance regions.

These frontotemporal regions are consistent with the \textit{language network} areas identified with individual-subject localizers in \cite{fedorenko2010new} and by sensitivity to sentence constituent structure in \cite{pallier2011cortical}.
This method only describes model sensitivity and does not establish causal relationship between these regions and the decoding task.
Nonetheless, the prominence of such plausible regions across the models is still encouraging given that they were trained without any prior on language-responsive regions.

Also, those maps do not show a clear left-right asymmetry that could be expected given the well-established left lateralization of language processing \citep{broca1865siege}.
However, while linearly aligning language-model representations with fMRI responses, \cite{bonnasse2024fmri} found that
this asymmetry re-appears for larger language models.
Therefore, despite the different setup, larger scale models might be necessary for the same re-emergence to occur.

\section{Conclusion}
\label{sec:conc}

NOEMA combines temporal modeling of fMRI with alignment to text, audio, and
speech representations to improve naturalistic speech retrieval. On entirely
held-out recording sessions containing speech stimuli absent from training,
NOEMA outperforms all baselines across two naturalistic listening datasets,
demonstrating generalization to new sessions and stimuli within participants.
This advantage is preserved when the retrieval set is extended with external
distractors.

Our ablations show that both temporal context and the choice of target
representations contribute to retrieval performance, supporting the view that
the temporal structure of brain measurements and the representations used for
supervision are complementary design choices for speech decoders. NOEMA
benefits more from additional fMRI frames than an MLP receiving the same
context, despite having fewer parameters, and its performance depends on
chronological frame order.

\textbf{Limitations and future work.}
Our evaluation is limited to six participants across two listening datasets, with a separate decoder trained for each participant; generalization to new individuals and to other tasks remains untested, and extending NOEMA across participants is a natural next step. Our analyses also do not establish what underlies the benefit of temporal context: it may reflect accommodation of the delayed hemodynamic response, stimulus-related temporal structure, or both. Moreover, the temporal window is bounded by the architecture's seven-frame receptive field, and the current results do not establish an optimal context length. Finally, NOEMA retrieves from a predefined candidate set. Although its advantage persists with external distractors, retrieval from any finite set does not demonstrate unrestricted speech reconstruction; using NOEMA's predicted embeddings to guide text generation is a promising direction toward this goal.

\clearpage
\subsection*{AI use statement}

We used generative AI tools extensively throughout the research and writing process.
Their use included developing and debugging code, preparing and running experiments, analyzing and interpreting results, producing figures, searching the literature, and drafting and revising the manuscript.
Coauthors reviewed AI-assisted contributions through a pull-request process to check their correctness.
The authors take full responsibility for the methods, results, and final content of this work.

\subsection*{Ethics statement}

This paper presents a methodological study of naturalistic speech decoding from fMRI using two existing datasets: \textit{Cephalonauts One} \citep{collas2026cephalonauts} and \textit{LeBel} \citep{lebel_natural_2023}.
Both datasets were collected under protocols approved by the respective ethics committees, and all subjects provided written informed consent.

\subsection*{Reproducibility statement}

We describe the datasets, preprocessing, data splits, training procedure, and evaluation protocol in Section~\ref{sec:experimental_setup}.
Where available, we use the test sessions specified by the dataset benchmark, as for \textit{Cephalonauts One}~\citep{collas2026cephalonauts}.
Appendix~\ref{app:detailed_configurations} provides the detailed architecture, a standalone PyTorch implementation of NOEMA, and the hyperparameter search space and selected configurations.

\subsubsection*{Acknowledgments}
We thank the developers and maintainers of the open-source software used in this work.
Our Python implementation uses NumPy~\citep{harris2020array}, SciPy~\citep{virtanen_scipy_2020}, pandas~\citep{mckinney2010data}, xarray~\citep{hoyer2017xarray}, scikit-learn~\citep{pedregosa_scikit-learn_2011}, and chardet for data handling and analysis.
Neuroimaging preprocessing uses DeepPrep~\citep{ren2025deepprep}, Nilearn~\citep{abraham_machine_2014}, and NiBabel, while audio processing uses TorchAudio~\citep{yang2022torchaudio}, librosa~\citep{mcfee2015librosa}, and SoundFile.
Model training and stimulus embedding extraction rely on PyTorch~\citep{paszke2019pytorch}, Transformers~\citep{wolf_transformers_2020}, Sentence Transformers~\citep{reimers2019sentence}, and Hugging Face Hub.
We use Optuna~\citep{akiba2019optuna}, Exca, Submitit, joblib, Pydantic, and PyYAML to configure, optimize, and run experiments, and MLflow~\citep{zaharia2018mlflow}, Loguru, psutil, and tqdm~\citep{da_costa_luis2019tqdm} to monitor them.
Visualizations use Matplotlib~\citep{hunter_matplotlib_2007} and Plotly.

\bibliography{noema}
\bibliographystyle{plainnat}

\clearpage
\appendix

\section{Dataset size}
\label{app:retained_data}

\begin{table}[htbp]
    \centering
    \caption{\textbf{Number of sessions, runs, and hours per participant.}
    Counts cover all training, validation, and test sessions, with each run counted once.
    LeBel localizer runs and repeated presentations of \emph{Where There's Smoke} are excluded.
    Hours are computed from the number of native fMRI frames multiplied by the TR, before temporal downsampling and window selection.}
    \label{tab:retained_data}
    \begin{tabular}{llrrr}
        \toprule
        Dataset & Participant & Sessions & Runs & Hours \\
        \midrule
        Cephalonauts One & sub-00 & 24 & 120 & 29.94 \\
                        & sub-01 & 24 & 121 & 30.24 \\
                        & sub-02 & 24 & 117 & 29.13 \\
        \midrule
        LeBel           & UTS01 & 15 & 82 & 16.14 \\
                        & UTS02 & 15 & 83 & 16.37 \\
                        & UTS03 & 15 & 83 & 16.37 \\
        \bottomrule
    \end{tabular}
\end{table}

\section{Detailed configurations}
\label{app:detailed_configurations}

\subsection{Model architecture}
\label{app:noema_architecture}
Table~\ref{tab:noema_architecture} lists each operation of the NOEMA decoder with its output shape, parameter count, and per-example compute.

\begin{table}[htbp]
    \centering
    \caption{\textbf{NOEMA decoder architecture.}
    Shapes are time $\times$ features, with $N=7$ frames, $V=20{,}484$ cortical vertices, $H=4096$, and $D=8704$.
    $\operatorname{Conv}_{K,d,C}^{(\ell)}$ is a temporal convolution with kernel size $K$, dilation $d$, and $C$ output channels, in TCN block $\ell$ (blank for the spatial projection); $\operatorname{Linear}_{C}$ is a linear layer with $C$ output channels.
    Parameter counts and per-example compute are given as orders of magnitude, omitting constants and biases.}
    \label{tab:noema_architecture}
    \small
    \setlength{\tabcolsep}{3pt}
    \renewcommand{\arraystretch}{1.15}
    \begin{tabular}{@{}llll@{}}
        \toprule
        Operation & Shape & Params & Compute \\
        \midrule
        Input window & $N\times V$ & -- & -- \\
        $\operatorname{Conv}_{1,1,H}$ & $N\times H$ & $VH$ & $NVH$ \\
        GELU; dropout $p=0.7$ & $N\times H$ & -- & $NH$ \\
        \midrule
        $\operatorname{Conv}_{3,1,H}^{(1)}$ & $N\times H$ & $H^2$ & $NH^2$ \\
        GELU; $\operatorname{Conv}_{1,1,H}^{(1)}$ & $N\times H$ & $H^2$ & $NH^2$ \\
        Residual; LayerNorm & $N\times H$ & $H$ & $NH$ \\
        \midrule
        $\operatorname{Conv}_{3,2,H}^{(2)}$ & $N\times H$ & $H^2$ & $NH^2$ \\
        GELU; $\operatorname{Conv}_{1,1,H}^{(2)}$ & $N\times H$ & $H^2$ & $NH^2$ \\
        Residual; LayerNorm & $N\times H$ & $H$ & $NH$ \\
        \midrule
        Select central frame & $H$ & -- & -- \\
        $\operatorname{Linear}_{D}$ & $D$ & $HD$ & $HD$ \\
        \midrule
        Total & -- & \shortstack[l]{$VH+H^2$\\$+HD$} & \shortstack[l]{$NVH+NH^2$\\$+HD$} \\
        \bottomrule
    \end{tabular}
\end{table}

\clearpage
Figure~\ref{fig:noema_code} gives the full model definition in PyTorch~\citep{paszke2019pytorch}: a spatial projection applied independently to each frame, two TCN blocks, and a linear layer applied to the central frame.

\definecolor{githubtext}{HTML}{24292E}
\definecolor{githubkeyword}{HTML}{D73A49}
\definecolor{githubcomment}{HTML}{6A737D}
\definecolor{githubstring}{HTML}{032F62}
\definecolor{githubentity}{HTML}{6F42C1}
\definecolor{githubbg}{HTML}{FFFFFF}

\begin{figure}[H]
\begin{lstlisting}[
    language=Python,
    basicstyle=\ttfamily\footnotesize\color{githubtext},
    keywordstyle=\color{githubkeyword},
    commentstyle=\color{githubcomment}\itshape,
    stringstyle=\color{githubstring},
    emph={NOEMA,TCNblock,Module,Sequential,Conv1d,Linear,LayerNorm,GELU,Dropout},
    emphstyle=\color{githubentity},
    backgroundcolor=\color{githubbg},
    columns=fullflexible,
    keepspaces=true,
    showstringspaces=false,
    frame=none
]
import torch.nn as nn


class TCNblock(nn.Module):
    def __init__(self, hidden_dim, dilation):
        super().__init__()
        self.layers = nn.Sequential(
            nn.Conv1d(hidden_dim, hidden_dim, kernel_size=3,
                      padding=dilation, dilation=dilation),
            nn.GELU(),
            nn.Conv1d(hidden_dim, hidden_dim, kernel_size=1),
        )
        self.norm = nn.LayerNorm(hidden_dim)

    def forward(self, x):
        # x: (batch, frames, hidden_dim)
        return self.norm(
            x + self.layers(x.transpose(1, 2)).transpose(1, 2)
        )


class NOEMA(nn.Module):
    def __init__(self, input_dim, output_dim=8704,
                 hidden_dim=4096, dropout=0.7):
        super().__init__()
        self.spatial = nn.Sequential(
            nn.Linear(input_dim, hidden_dim),
            nn.GELU(),
            nn.Dropout(dropout),
        )
        self.temporal = nn.Sequential(
            TCNblock(hidden_dim, dilation=1),
            TCNblock(hidden_dim, dilation=2),
        )
        self.output = nn.Linear(hidden_dim, output_dim)

    def forward(self, x):
        # x: (batch, frames, input_dim).
        x = self.temporal(self.spatial(x))

        # project the central frame
        return self.output(x[:, x.shape[1] // 2])
\end{lstlisting}
\caption{\textbf{PyTorch implementation of NOEMA.}
The spatial projection is applied independently to each frame, and two TCN blocks with dilations 1 and 2 give the central frame a seven-frame receptive field.
Only the central frame is projected to the target space.
Default dimensions follow Table~\ref{tab:noema_architecture}.}
\label{fig:noema_code}
\end{figure}

\clearpage
\subsection{Hyperparameter Optimization}
\label{app:hpo}

We use Optuna~\citep{akiba2019optuna} to optimize the training hyperparameters for NOEMA and the baselines.
Specifically, we perform a Bayesian search over the learning rate, weight decay, batch size, temperature, and hidden width of the neural models, and over the penalty $\alpha$ of the Ridge decoder.
Other architecture choices, dropout rates, and early stopping patience are fixed to the values used in the experiments; for MindEye-MLP, dropout rates are the default values from \cite{scotti_reconstructing_2023}.
Test sessions are excluded entirely from this search.
We use five session-based training/validation splits of the remaining sessions, each holding out one session for validation and using the others for training.
The search is performed only on \textit{Cephalonauts One} subject~\texttt{sub-01}, using the concatenated text--audio--speech target representation.
The resulting search space is summarized in Table~\ref{tab:hyperparams}.
The best hyperparameters found for each model are reported in Table~\ref{tab:best_hyperparams}.

\begin{table}[h]
\centering
\caption{\textbf{Hyperparameter search space.}
Bayesian search with Optuna on \textit{Cephalonauts One} sub-01, using five session-based training/validation splits of the non-test sessions.
Each trial minimizes the validation median relative rank averaged over folds.
Neural models are NOEMA, both MLP variants, and MindEye-MLP; hyperparameters not listed are fixed to the values in Section~\ref{sec:training}.}
\label{tab:hyperparams}
\begin{tabular}{llll}
\toprule
Model & Hyperparameter & Search space & Sampling \\
\midrule
\multirow{5}{*}{Neural models} & Batch size & $\{2048, 4096, 8192\}$ & categorical \\
 & Hidden width & $\{1024, 2048, 4096, 8192\}$ & categorical \\
 & Temperature & $[10^{-3},\,1]$ & log-uniform \\
 & Learning rate & $[10^{-5},\,10^{-2}]$ & log-uniform \\
 & Weight decay & $[10^{-5},\,10^{-2}]$ & log-uniform \\
\midrule
Ridge & Penalty $\alpha$ & $[10^{-1},\,10^{5}]$ & log-uniform \\
\bottomrule
\end{tabular}
\end{table}

\begin{table}[h]
\centering
\caption{\textbf{Selected hyperparameters.}
Best Optuna trial for each decoder, from the search space in Table~\ref{tab:hyperparams}.
The last column gives the validation median relative rank (mean $\pm$ standard deviation over five folds; lower is better).
Ridge uses $\alpha=4.50\times10^{3}$.
Experiments use temperature, learning rate, weight decay, and Ridge penalty rounded to two significant figures; batch sizes and hidden widths are used exactly as listed.}
\label{tab:best_hyperparams}
\small
\setlength{\tabcolsep}{4pt}
\begin{tabular}{@{}lcccccc@{}}
\toprule
Model & Batch size & Width & $\tau$ & Learning rate & Weight decay & Val.\ rel.\ rank $\downarrow$ \\
\midrule
NOEMA          & 4096 & 4096 & $6.23\times10^{-3}$ & $1.13\times10^{-4}$ & $6.79\times10^{-3}$ & $0.0077 \pm 0.0024$ \\
Context MLP & 4096 & 4096 & $8.79\times10^{-3}$ & $3.68\times10^{-5}$ & $5.33\times10^{-3}$ & $0.0128 \pm 0.0029$ \\
MLP  & 8192 & 8192 & $9.12\times10^{-3}$ & $1.30\times10^{-4}$ & $4.26\times10^{-5}$ & $0.0193 \pm 0.0047$ \\
MindEye-MLP    & 4096 & 2048 & $8.83\times10^{-3}$ & $6.65\times10^{-4}$ & $2.95\times10^{-5}$ & $0.0203 \pm 0.0046$ \\
Ridge          & --   & --   & --                  & --                  & --                  & $0.0919 \pm 0.0186$ \\
\bottomrule
\end{tabular}
\end{table}

\clearpage
\section{Sensitivity to lag}
\label{app:lag_ablation}

\begin{figure}[H]
    \centering
    \includegraphics[width=\linewidth]{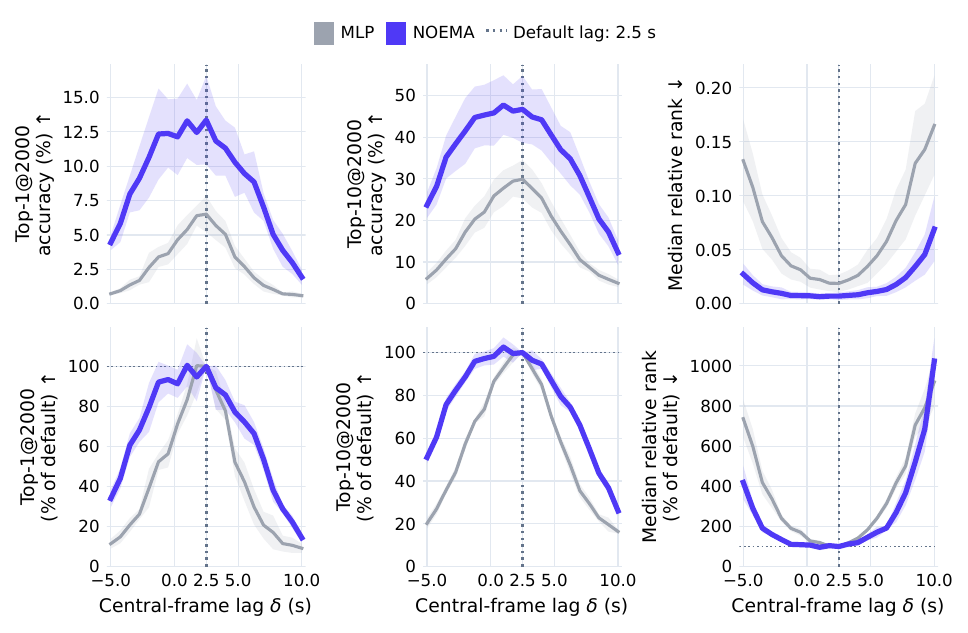}
    \caption{\textbf{Sensitivity to the alignment offset (\textit{Cephalonauts One}, sub-01).}
Columns: Top-1@2000 and Top-10@2000 accuracy (higher is better), and median relative rank (lower is better).
Top row: absolute values. Bottom row: values as a percentage of each model's value at the default $\delta=2.5$ seconds (dotted lines), computed within each fold before averaging.
Lines show the mean over five folds; shaded bands show one standard deviation.}
    \label{fig:lag_ablation}
\end{figure}

We vary the alignment offset $\delta$ (Section~\ref{sec:architecture}) from $-5$ to $10$ seconds in $0.75$-second steps on \textit{Cephalonauts One} sub-01, comparing NOEMA with the MLP.
Both models receive the same seven-frame windows, but the MLP uses only the central frame.
Each model is retrained at every offset on five session folds, for 210 fits in total.
NOEMA is more accurate than the MLP at every offset (Figure~\ref{fig:lag_ablation}) and is also less sensitive to the choice of offset: its Top-10@2000 accuracy stays within 10\% of its value at the default $\delta=2.5$ seconds for offsets from $-1.25$ to $4$ seconds, compared with $1$ to $3.25$ seconds for the MLP.
This robustness is expected, since NOEMA's receptive field spans approximately $\pm6$ seconds around the central frame.
NOEMA's accuracy peaks at $\delta=1$ second (47.75\%), 1.04 percentage points above the default (46.71\%), whereas the MLP peaks at the default (29.90\%).
Because these curves come from the test sessions of a single participant, we do not use them to select $\delta$.
The offset is measured from the end of the stimulus context, so it is not an estimate of the hemodynamic response peak.

\clearpage
\section{ASR context window}
\label{app:asr_context_ablation}
\begin{figure}[H]
    \centering
    \includegraphics[width=\linewidth]{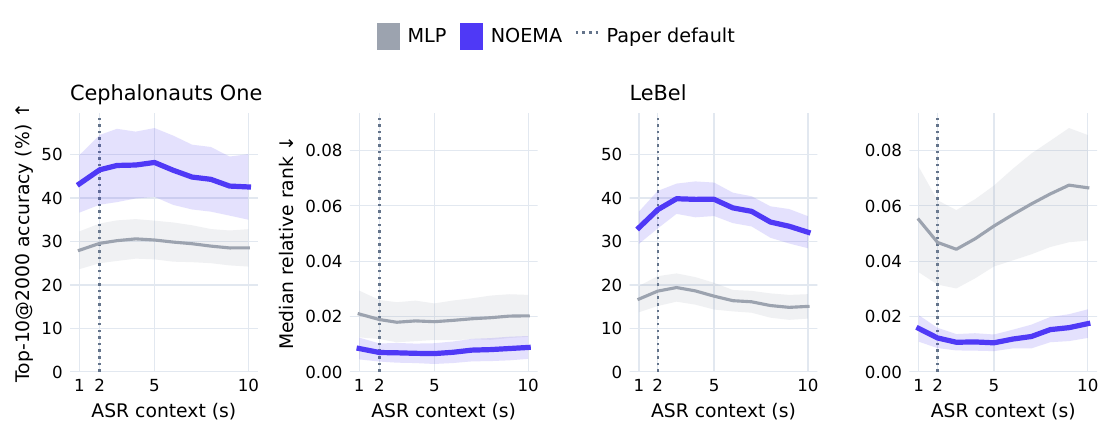}
    \caption{\textbf{Effect of the ASR context window.}
Top-10@2000 accuracy (higher is better) and median relative rank (lower is better) as a function of the trailing Parakeet window, for \textit{Cephalonauts One} sub-01 (left pair) and \textit{LeBel} UTS01 (right pair).
Dotted lines mark the default window, equal to the marker spacing $\Delta$ (2.08 and 2.0 seconds).
Lines show the mean over five folds; shaded bands show one standard deviation.}
    \label{fig:asr_context_ablation}
\end{figure}

Section~\ref{sec:stimuli} sets the trailing context window of the ASR encoder equal to the marker spacing $\Delta$ (2.08 seconds for \textit{Cephalonauts One}, 2.0 seconds for \textit{LeBel}), so that successive speech targets tile the stimulus without overlap.
Here we test whether a different window improves retrieval.
We vary the Parakeet window over 1 second, $\Delta$, and 3 to 10 seconds in 1-second steps, keeping the marker spacing $\Delta$ and the text and audio context $T=10$ seconds fixed.
NOEMA and the MLP are retrained at each window with the hyperparameters of Appendix~\ref{app:hpo} on five session folds, for one subject per dataset (\textit{Cephalonauts One} sub-01 and \textit{LeBel} UTS01), for 200 fits in total.
Accuracy follows an inverted-U shape for both models on both datasets (Figure~\ref{fig:asr_context_ablation}).
NOEMA's Top-10@2000 accuracy peaks at 5 seconds on \textit{Cephalonauts One} (48.19\%, compared with 43.27\% at 1 second and 42.55\% at 10 seconds) and at 3 seconds on \textit{LeBel} (39.84\%, compared with 33.18\% and 32.13\%).
The default window lies 1.70 and 2.52 percentage points below these peaks, respectively; for the MLP, the gaps are 1.03 and 0.82 points.
Validation accuracy peaks at the same windows, so tuning the ASR window per dataset could bring further gains.
We keep the window equal to $\Delta$ because it gives each target a distinct, non-overlapping speech segment, and leave this tuning to future work.

\clearpage
\section{Evaluation with additional metrics}
\label{app:additional_results}

\subsection{Top-1 accuracy and median relative rank}
\label{app:top1_median_rank}
\begin{figure}[H]
    \centering
    \includegraphics[width=\linewidth]{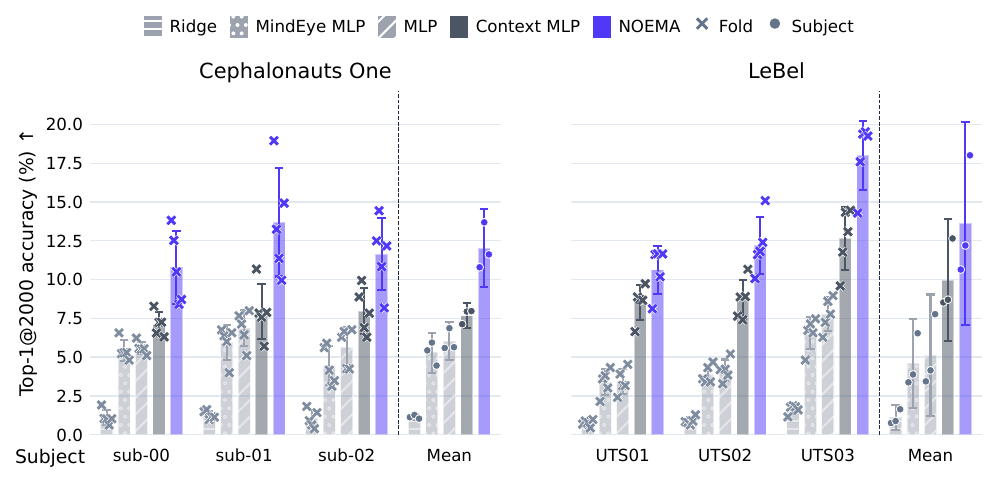}
    \includegraphics[width=\linewidth]{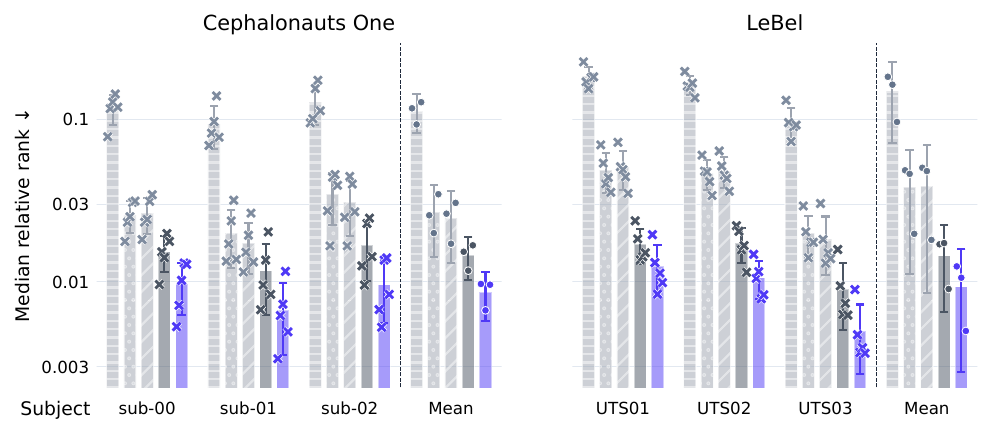}
    \caption{\textbf{Top-1@2000 accuracy and median relative rank.}
\textbf{Top:}~Top-1@2000 accuracy (higher is better).
\textbf{Bottom:}~Median relative rank of the correct segment, on a log scale (lower is better).
Models and colors follow Figure~\ref{fig:retrieval_comparison}.
In subject columns, bars and error bars show the mean and standard deviation over five folds (crosses); in Mean columns, they show the mean and 90\% Student-$t$ confidence interval over three subjects (circles).}
    \label{fig:additional_metrics}
\end{figure}

Figure~\ref{fig:additional_metrics} complements the Top-10@2000 accuracy in Figure~\ref{fig:retrieval_comparison} with two further metrics.
Top-1@2000 accuracy is the strictest criterion: the correct segment must rank first among approximately 2,000 candidates.
Median relative rank instead summarizes the typical position of the correct segment divided by the retrieval-set size, including queries where it falls outside the top ten.
NOEMA performs best on both metrics for every subject.
It ranks the correct segment first for 12.0\% of queries on \textit{Cephalonauts One} and 13.6\% on \textit{LeBel}, compared with 7.7\% and 10.0\% for the strongest baseline, the seven-frame MLP.
It also reduces the mean median relative rank from 0.0146 to 0.0087 on \textit{Cephalonauts One} and from 0.0145 to 0.0093 on \textit{LeBel}.

\clearpage
\subsection{Rank distributions}
\label{app:rank_dists}
\begin{figure}[H]
    \centering
    \includegraphics[width=\linewidth]{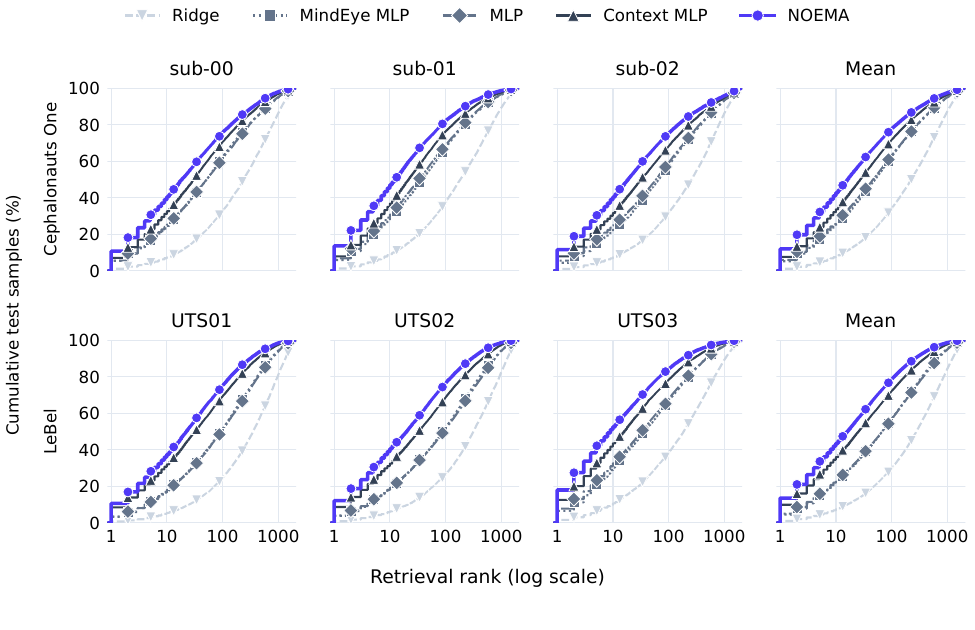}
    \caption{\textbf{Cumulative distribution of retrieval ranks.}
Percentage of test queries whose correct segment ranks at or above a given position, for the models in Figure~\ref{fig:retrieval_comparison}; ranks are on a log scale.
Values at ranks 1 and 10 equal Top-1@2000 and Top-10@2000 accuracy.
Rows show datasets; columns show individual subjects and the mean over subjects, with folds weighted equally.
Higher curves indicate better retrieval.}
    \label{fig:rank_cdf}
\end{figure}

\begin{figure}[H]
    \centering
    \includegraphics[width=\linewidth]{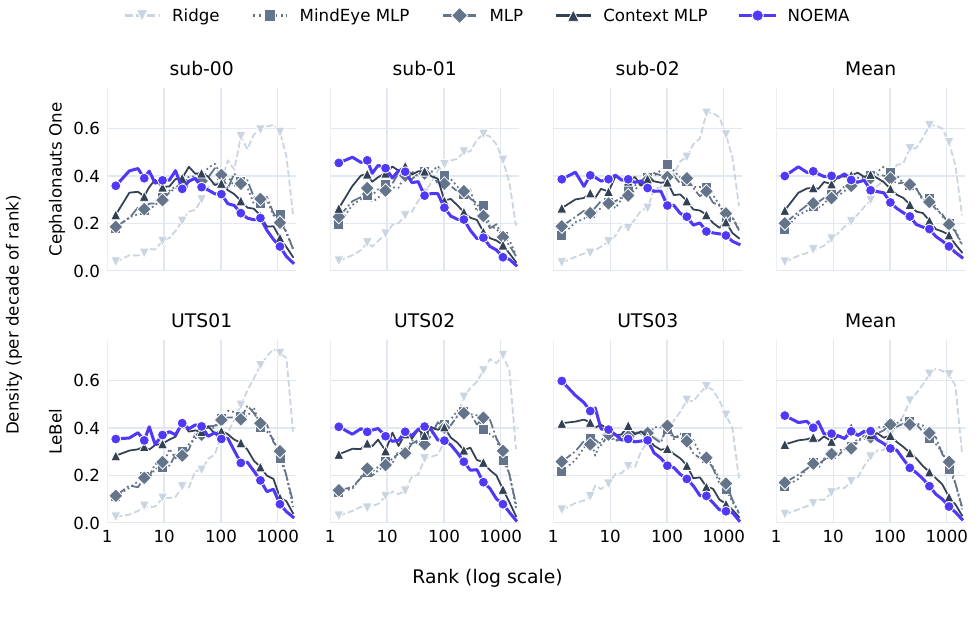}
    \caption{\textbf{Distribution of retrieval ranks.}
Density of the correct segment's rank for the models in Figure~\ref{fig:retrieval_comparison}, with ranks binned on a log scale (rank $k$ falls in $[k, k{+}1)$).
Density is per decade of rank, so the area under each curve over $\log_{10}$ rank is one.
Panels follow Figure~\ref{fig:rank_cdf}.
Curves shifted to the left indicate better retrieval.}
    \label{fig:rank_histograms}
\end{figure}

Top-$k$ accuracy and median rank each summarize the retrieval ranks with a single number.
Figures~\ref{fig:rank_cdf} and~\ref{fig:rank_histograms} show the full distribution of the correct segment's rank, for the models in Figure~\ref{fig:retrieval_comparison}.
The cumulative distributions show that NOEMA's advantage is not limited to a particular threshold. For every subject, its curve lies on or above that of every baseline across nearly the entire rank range, with small crossings as all models approach 100\%.
Averaged over subjects, NOEMA places half of all queries within rank 16 on both datasets, compared with ranks 28 and 25 for the seven-frame MLP on \textit{Cephalonauts One} and \textit{LeBel}, respectively.
It places 90\% of queries within rank 328 on \textit{Cephalonauts One} and 263 on \textit{LeBel}, compared with 445 and 390.

\clearpage
The histograms show where this advantage comes from.
NOEMA's density is highest among the lowest ranks and decreases steadily beyond rank~100, so most of its queries fall within the top few percent of the approximately 2,000 candidates.
The baselines instead peak at intermediate ranks, around 100 for the MLPs, and Ridge places most of its mass near the top of the range, where a decoder that ranked candidates at random would also place most of its mass.
Moving from Ridge to the MLPs, the seven-frame MLP, and NOEMA, probability mass shifts steadily toward lower ranks, the same ordering as in Figure~\ref{fig:retrieval_comparison}.

\clearpage
\section{stimulus representations and temporal context}

\subsection{Results on \textit{LeBel} dataset}
\label{app:lebel_modality_temporal_ablation}
\begin{figure}[H]
    \centering
    \includegraphics[width=\linewidth]{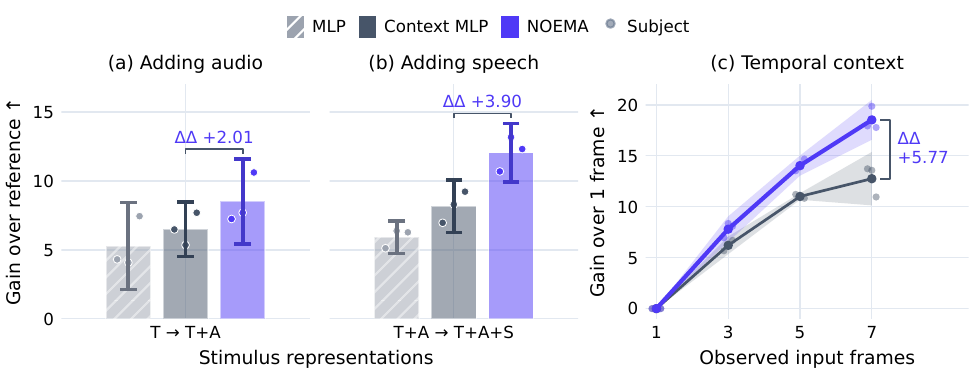}
    \caption{\textbf{Effect of stimulus representations and temporal context (\textit{LeBel}).}
Gain in Top-10@2000 accuracy (percentage points) over a reference configuration, computed within each model, subject, and fold.
\textbf{(a)}~Adding audio to a text-only target (T$\rightarrow$T+A).
\textbf{(b)}~Adding speech to a text-and-audio target (T+A$\rightarrow$T+A+S).
\textbf{(c)}~Increasing the input window from one frame to three, five, and seven frames (Context MLP and NOEMA only).
T: Qwen text; A: LCO audio; S: Parakeet speech.
$\Delta\Delta$ is the mean NOEMA gain minus the mean Context MLP gain, at seven frames in (c).
Circles show subject means over five folds; bars in (a, b) and lines in (c) show the mean over three subjects, with 90\% Student-$t$ confidence intervals.}
    \label{fig:modality_temporal_ablation_lebel}
\end{figure}

Figure~\ref{fig:modality_temporal_ablation_lebel} repeats the experiments of Figure~\ref{fig:modality_temporal_ablation} on \textit{LeBel}.
Adding audio to a text-only target improves NOEMA's Top-10@2000 accuracy by 8.52 percentage points, and adding speech improves it by a further 12.06 points, exceeding the Context MLP's gains by 2.01 and 3.90 points, respectively.
The speech addition thus has a larger effect on \textit{LeBel} than on \textit{Cephalonauts One} (7.25 points).
Extending the input window from one to seven frames improves NOEMA by 18.53 points and the Context MLP by 12.76 points; unlike on \textit{Cephalonauts One}, the Context MLP keeps improving up to seven frames.

\subsection{Absolute retrieval accuracies across stimulus representations}
\label{app:abs_stim_rep}
\begin{figure}[H]
    \centering
    \includegraphics[width=\linewidth]{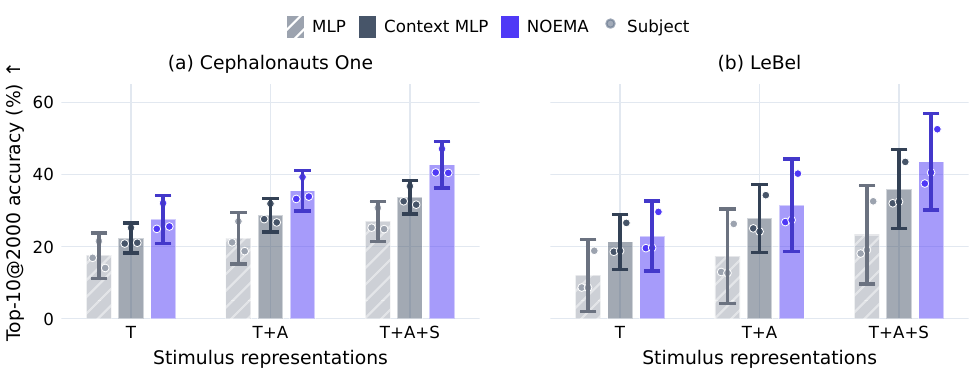}
    \caption{\textbf{Top-10@2000 accuracy across stimulus representations.}
Absolute accuracies underlying panels (a) and (b) of Figures~\ref{fig:modality_temporal_ablation} and~\ref{fig:modality_temporal_ablation_lebel}, for \textbf{(a)}~\textit{Cephalonauts One} and \textbf{(b)}~\textit{LeBel}.
T: text; A: audio; S: speech.
MLP uses one input frame; Context MLP and NOEMA use seven.
Circles show subject means over five folds; bars and error bars show the mean and 90\% Student-$t$ confidence interval over three subjects.}
    \label{fig:modality_ablation_incremental_absolute}
\end{figure}

Figure~\ref{fig:modality_ablation_incremental_absolute} reports the absolute accuracies underlying the gains in Figures~\ref{fig:modality_temporal_ablation} and~\ref{fig:modality_temporal_ablation_lebel}.
NOEMA achieves the highest mean accuracy for every target combination on both datasets, and every model is most accurate with the full T+A+S target.

\subsection{Absolute retrieval accuracies with increasing temporal context}
\label{app:abs_temporal_context}
\begin{figure}[H]
    \centering
    \includegraphics[width=\linewidth]{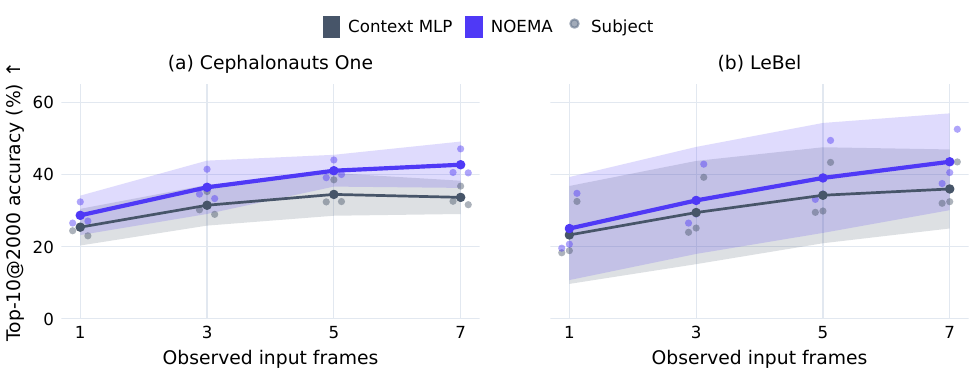}
    \caption{\textbf{Top-10@2000 accuracy with increasing temporal context.}
Absolute accuracies underlying panel (c) of Figures~\ref{fig:modality_temporal_ablation} and~\ref{fig:modality_temporal_ablation_lebel}, for \textbf{(a)}~\textit{Cephalonauts One} and \textbf{(b)}~\textit{LeBel}.
Circles show subject means over five folds; lines and shaded bands show the mean and 90\% Student-$t$ confidence interval over three subjects.
The wide bands on \textit{LeBel} reflect differences in overall accuracy between subjects; the within-subject gains in Figure~\ref{fig:modality_temporal_ablation_lebel}c are the more direct comparison.}
    \label{fig:temporal_context_absolute}
\end{figure}

Figure~\ref{fig:temporal_context_absolute} reports the absolute accuracies underlying the temporal-context gains.
On average, NOEMA is more accurate than the Context MLP at every window length on both datasets, and the gap widens with more frames: from 3.3 to 9.1 percentage points on \textit{Cephalonauts One} and from 1.8 to 7.6 points on \textit{LeBel} between one and seven frames.

\clearpage
\subsection{Retrieval accuracy and model size}
\label{app:model_size}
\begin{figure}[H]
    \centering
    \includegraphics[width=\linewidth]{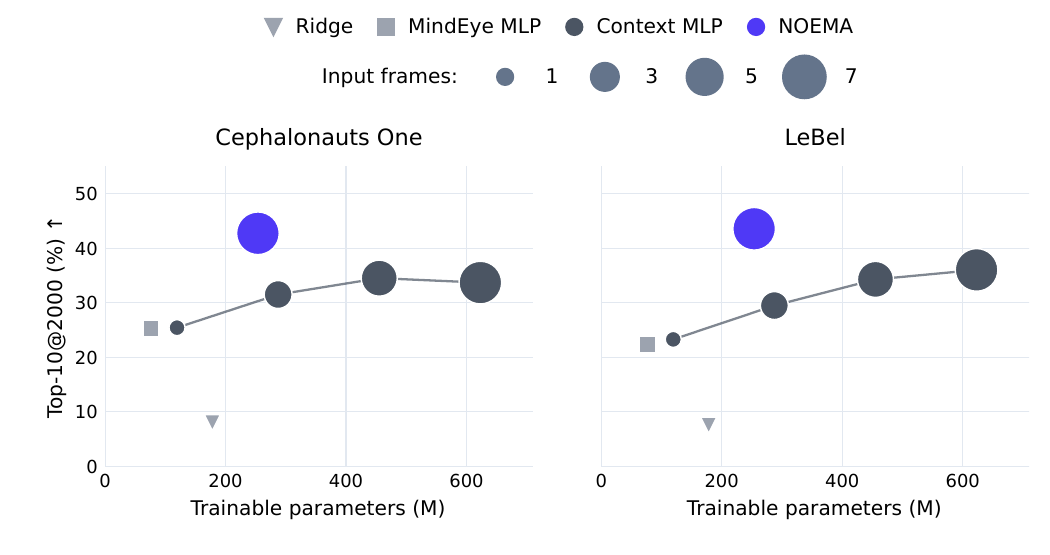}
    \caption{\textbf{Retrieval accuracy versus decoder parameter count.}
Context MLP uses one, three, five, or seven input frames; NOEMA is shown with seven frames only. MindEye MLP and Ridge use one frame.
Marker area is proportional to the number of input frames.
Points show mean Top-10@2000 accuracy across three subjects after averaging five folds per subject.
Counts include decoder weights, biases, and normalization parameters, and exclude frozen stimulus encoders.
All decoders use 20,484 input vertices and 8,704 output features. Hidden widths are 4,096 for Context MLP and NOEMA, and 2,048 for MindEye MLP.}
    \label{fig:model_size}
\end{figure}

Figure~\ref{fig:model_size} relates retrieval accuracy to the number of decoder parameters.
Context MLP grows from 119.6M to 623.0M parameters as its input increases from one to seven frames.
NOEMA uses 253.8M parameters independently of the input window length; its seven-frame configuration achieves higher mean accuracy than all four Context MLP configurations on both datasets.
This comparison shows that NOEMA's advantage does not come from a larger parameter count.

\clearpage
\section{Sensitivity to fMRI frame order}
\label{app:brain_context_shuffle}

\begin{figure}[H]
    \centering
    \includegraphics[width=0.85\linewidth]{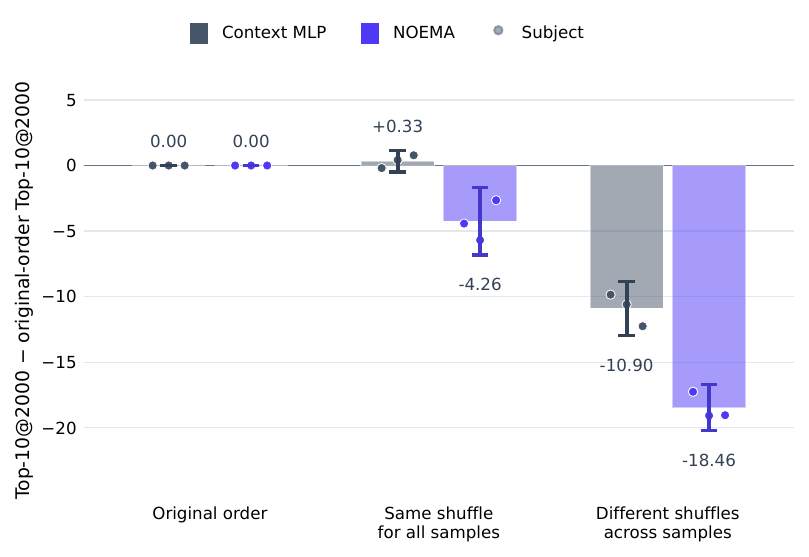}
    \caption{\textbf{Sensitivity of Context MLP and NOEMA to fMRI frame order on \textit{Cephalonauts One}.}
    We randomly permute the temporal dimension of each seven-frame fMRI input.
    Global shuffling applies a single fixed permutation to all samples across subjects, folds, and data splits.
    Per-sample shuffling assigns each sample its own permutation, fixed across epochs.
    Context MLP and NOEMA are trained from scratch for each condition, with the same shuffling rule applied during training, validation, and testing.
    We report the change in Top-10@2000 accuracy relative to training and evaluating on the original frame order, paired by model, subject, and fold.
    Each circle represents one subject: paired differences are first averaged over available folds within each permutation seed, then equally across permutation seeds within the subject.
    Bars show means across three subjects, with 90\% Student-$t$ confidence intervals.}
    \label{fig:brain_context_shuffle}
\end{figure}

Figure~\ref{fig:brain_context_shuffle} compares seven-frame Context MLP and NOEMA after retraining and testing with shuffled frame orders.
Each input retains exactly the same frame values and prediction target; only the assignment of frames to input positions changes.
Because the shuffling rule is applied during both training and evaluation, this experiment measures the benefit of the original temporal organization after adaptation, rather than robustness to a test-time perturbation.

\textbf{A shared permutation probes the architectural bias toward temporal order.}
Context MLP is nearly unchanged ($+0.33$ percentage points): its first dense layer operates on concatenated frames and can absorb any fixed input permutation by reordering its weights.
Thus, the model can learn position-specific temporal information without requiring neighboring input positions to correspond to neighboring acquisition times.
NOEMA instead loses $4.26$ points despite retraining.
Its local, shared temporal filters impose structure on how frames interact; arbitrary permutations alter temporal neighborhoods and their relation to the central readout.
This result suggests that NOEMA benefits from the chronological organization of fMRI frames, which its temporal inductive bias is designed to exploit.

\textbf{Per-sample permutations probe the value of consistent temporal positions.}
Shuffling each sample differently breaks the temporal ordering across samples.
Accuracy falls by $10.90$ points for Context MLP and $18.46$ for NOEMA.
Both models therefore benefit from consistent temporal positions, with a larger observed dependence for NOEMA.

\clearpage
\section{Training-data scaling}
\label{app:data_scaling}
We compare seven-frame Context MLP and NOEMA as the number of training sessions increases.
Both models use the same nested subsets of whole sessions, with validation and test sessions held fixed within each fold.
The six training budgets are 1, 2, 3, 6, 12, and 22 sessions for \textit{Cephalonauts One}, and 1, 2, 3, 5, 8, and 13 for \textit{LeBel}.

\begin{figure}[H]
    \centering
    \includegraphics[width=\linewidth]{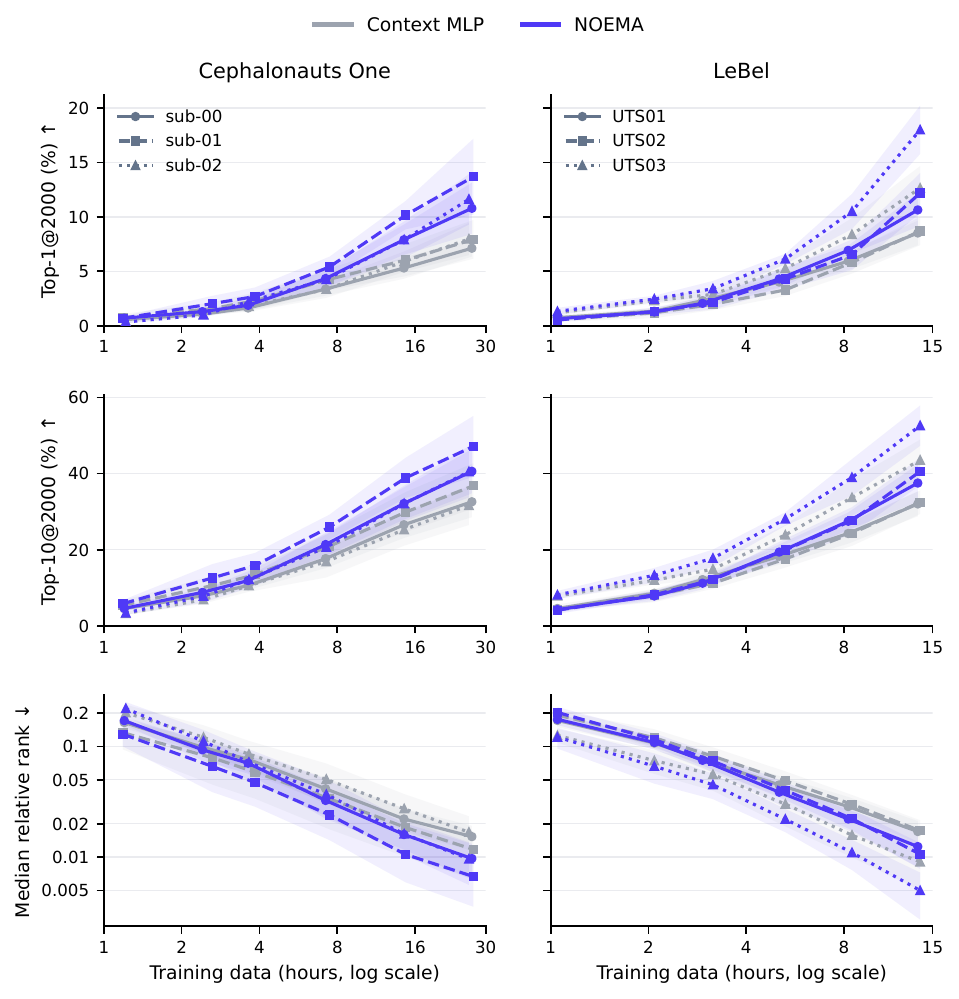}
    \caption{\textbf{Retrieval performance versus training-data duration.}
    Rows show Top-1@2000 accuracy, Top-10@2000 accuracy, and median relative rank (lower is better).
    Each curve represents one model and subject. Points average five session folds; shaded bands show $\pm1$ standard deviation across folds.
    Color identifies the model; line style and marker identify the subject.
    All horizontal axes and the median-relative-rank vertical axes are logarithmic.}
    \label{fig:scaling_top10}
    \label{fig:scaling_top1}
    \label{fig:scaling_rank}
\end{figure}

\clearpage
\section{Extended retrieval sets}
\label{app:extended_retrieval_set}
The extended retrieval sets in Section~\ref{sec:open_dictionary_results} draw on 50
additional episodes of the podcast show used in \textit{Cephalonauts One}, none of which appear
in the dataset.
We transcribe these episodes with the AssemblyAI \texttt{universal-3-pro} speech-to-text model,
as for the original stimuli~\citep{collas2026cephalonauts}, and split them into non-overlapping
ten-second segments.
Treating the end of each segment as a time marker $t$, we compute its text, audio, and speech
embeddings exactly as for the original stimuli (Section~\ref{sec:stimuli}), yielding 8,494
additional candidates.
For each test session, we add random subsets of 500, 1,000, 2,000, and 5,000 candidates, or all
8,494, to the original retrieval set of approximately 2,000 segments.
We evaluate the MLP, Context MLP, and NOEMA, trained as in
Section~\ref{sec:training}, on each extended set without retraining.

\section{Feature-importance maps}
\label{app:feature_importance}
\begin{figure}[H]
    \centering
    \includegraphics[width=\linewidth]{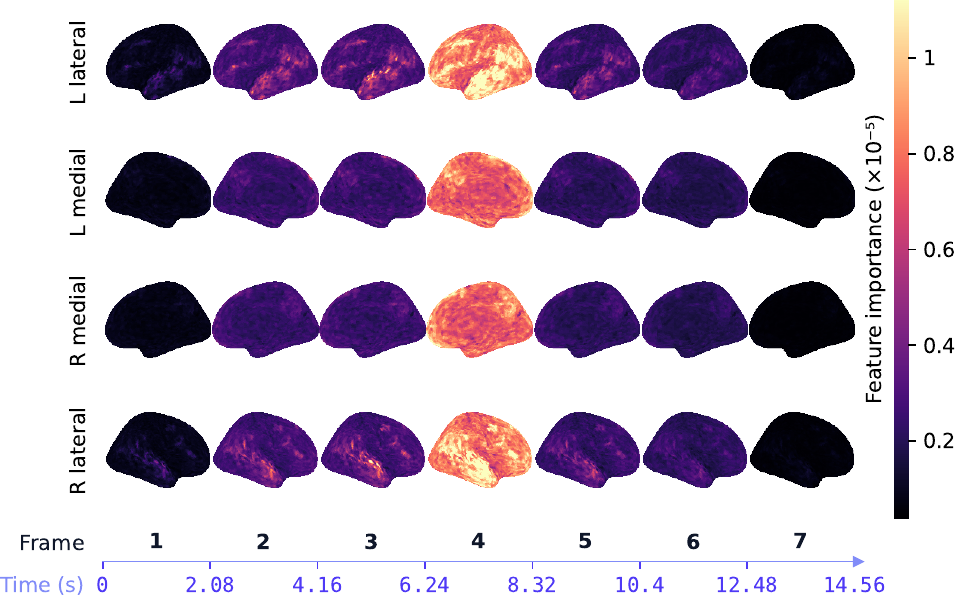}
    \caption{\textbf{NOEMA feature-importance maps.}
    Rows show the left lateral, left medial, right medial, and right lateral views and columns show frames 1--7.}
    \label{fig:feature_importance_all}
\end{figure}

\begin{figure}[H]
    \centering
    \includegraphics[width=\linewidth]{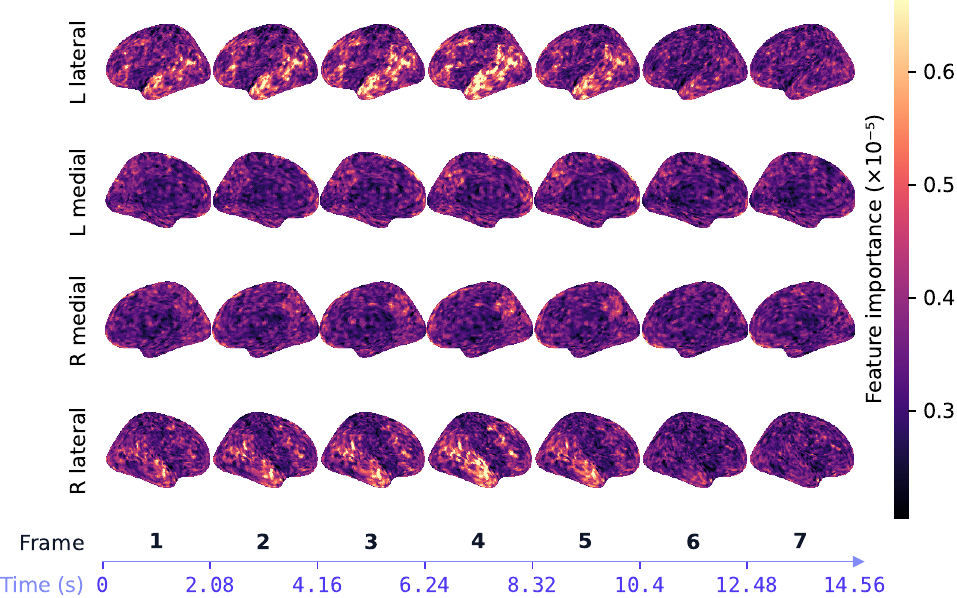}
    \caption{\textbf{Context MLP feature-importance maps.}
    Rows show the left lateral, left medial, right medial, and right lateral views and columns show frames 1--7.}
    \label{fig:feature_importance_mlp_7_frames}
\end{figure}

\begin{figure}[H]
    \centering
    \includegraphics[width=\linewidth]{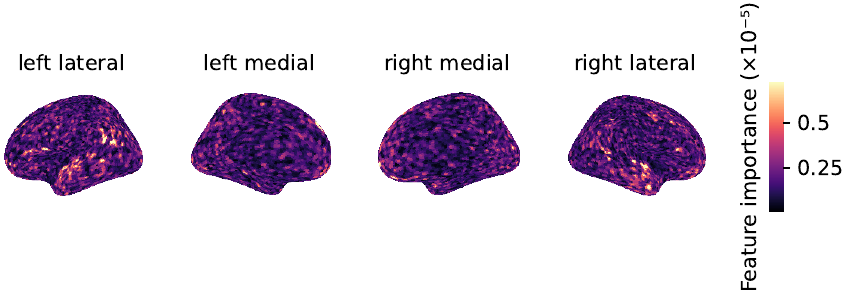}
    \caption{\textbf{One-frame MLP feature-importance maps.}
    From left to right: left lateral, left medial, right medial, and right lateral views.}
    \label{fig:feature_importance_mlp_1_frame}
\end{figure}

For the feature-importance analysis, we compute attributions for NOEMA, the one-frame MLP, and the seven-frame MLP on the held-out fold 0 test set of \textit{Cephalonauts One} \texttt{sub-01}.
We attribute each model's CLIP loss to every input vertex in each of its fMRI frames using 50-step integrated gradients from a zero baseline as implemented in Captum~\citep{sundararajan2017axiomatic,kokhlikyan2020captumunifiedgenericmodel}.
The choice of zero baseline is motivated by the fact that the input fMRI vertex time series are standardized within each run during preprocessing.
We average attributions over held-out samples and for better visualization we show absolute values of the attributions and the color scale spans from the 1st to the 99th percentiles over all the values.
\Cref{fig:feature_importance} in the main text shows NOEMA's lateral views, while \Cref{fig:feature_importance_all} adds the medial views.
The MLP maps appear in \Cref{fig:feature_importance_mlp_1_frame,fig:feature_importance_mlp_7_frames}.

\end{document}